\documentclass[journal]{IEEEtran}

\usepackage{graphicx}
\usepackage{cite}
\usepackage{amssymb}
\usepackage{latexsym,amscd}

\usepackage{graphicx}
\usepackage{float}
\usepackage{makecell}
\usepackage{times}

\usepackage{soul}
\usepackage{url}
\usepackage[hidelinks]{hyperref}
\usepackage[utf8]{inputenc}
\usepackage[small]{caption}
\usepackage{graphicx}
\usepackage{amsmath}
\usepackage{booktabs}
\usepackage{threeparttable}  
\usepackage{xcolor}

\usepackage{subfigure} 
\usepackage{multirow}
\usepackage{algorithm}
\usepackage{algorithmic}

\usepackage{ifthen}
\usepackage{float}
\usepackage{amssymb}

\usepackage{color} 
\ifCLASSINFOpdf
\else
\fi

\begin{document}

\title{HRR: Hierarchical Retrospection Refinement for Generated Image Detection}

\author{Peipei~Yuan, Zijing Xie, Shuo~Ye, Hong Chen, Yulong Wang% <-this % stops a space

\thanks{%This work was supported in part by the National Key R\&D Program (2022YFF0712300 and 2022YFC3301004). 
}

\thanks{Peipei~Yuan is with the School of Artificial Intelligence, Jianghan University, Wuhan, China.}
\thanks{Zijing~xie is with the School of Electronic Information and Communications, Huazhong University of Science and Technology, Wuhan, China.}
\thanks{Shuo~Ye is with the School of Electronic Information and Communications, Huazhong University of Science and Technology, Wuhan 430074, China. Corresponding author: \textit{Shuo Ye. e-mail: shuoye.ke@gmail.com.} }
\thanks{Hong Chen is with the College of Informatics, Huazhong Agricultural University, Wuhan 430070, China, and also with the Engineering Research Center of Intelligent Technology for Agriculture, Ministry of Education, Wuhan 430070, China}
\thanks{Yulong Wang is with the College of Informatics, Huazhong Agricultural University, Wuhan 430070, China}
}

%\markboth{IEEE Transactions on Circuits and Systems for Video Technology, 2023}%
%{Alignment of Multi-scale Features and Diverse Representations for Emotion Recognition}

\maketitle

\begin{abstract}
Generative artificial intelligence holds significant potential for abuse, and generative image detection has become a key focus of research. However, existing methods primarily focused on detecting a specific generative model and emphasizing the localization of synthetic regions, while neglecting the interference caused by image size and style on model learning. Our goal is to reach a fundamental conclusion: Is the image real or generated? To this end, we propose a diffusion model-based generative image detection framework termed Hierarchical Retrospection Refinement~(HRR). It designs a multi-scale style retrospection module that encourages the model to generate detailed and realistic multi-scale representations, while alleviating the learning biases introduced by dataset styles and generative models. Additionally, based on the principle of correntropy sparse additive machine, a feature refinement module is designed to reduce the impact of redundant features on learning and capture the intrinsic structure and patterns of the data, thereby improving the model's generalization ability. Extensive experiments demonstrate the HRR framework consistently delivers significant performance improvements, outperforming state-of-the-art methods in generated image detection task.
\end{abstract}

\begin{IEEEkeywords}
Generated image detection, Multi-scale features, Image decoupling, Additive models
\end{IEEEkeywords}

\section{Introduction}\label{introduction}
The maturity of deep learning techniques has significantly lowered the barrier for image forgery or manipulation. User-friendly image manipulation tools enable users to generate non-existent human faces or create convincing deepfakes~\cite{bai2025towards,hou2022guidedstyle,lee2024robust,lin2024detecting}. This development has brought about economic and legal challenges, and also threatens the integrity of social security, public health, and trust systems.

To address the issue of image forgery, the related research is divided into two main directions: image editing detection \cite{dong2022mvss, hu2020span, wang2022objectformer} and CNN-based synthesis detection~\cite{dang2020detection, wang2020cnn}. Editing detection typically trains a classifier on datasets composed of real images and operation images by synthetic techniques. Existing research is effective in detecting traces, such as image stitching~\cite{kwon2021cat}, artifacts from re-compression, resizing and other manipulations~\cite{corvi2023detection}. The CNN-based synthesis detection is commonly used to distinguish between naturally captured photographs and images fabricated by generative models, such as Generative Adversarial Networks~(GAN)~\cite{hussain2020high,goodfellow2014generative} and VAE~\cite{kingma2013auto}. However, the generalization of different generators is challenging, as the generative processes of different models (e.g., GAN and VAE) exhibit significant differences. This leads to the fact that  training a binary classifier on generated images from various generators and real images is not always reliable.

% Recent studies for generated image detection have revealed through spectral analysis that images generated by GANs inherently contain unique artifacts, exhibiting periodic, grid-like patterns in their spectra. 

Recent studies for generated image detection have shown that images generated by GANs inherently contain unique artifacts that exhibit periodic, grid-like patterns in the spectrum. 
These patterns significantly deviate from the spectral distribution of natural images, making them easier to identify~\cite{zhang2019detecting}. One of the key reasons for GANs' inability to replicate the spectral distribution of training data lies in the inherent use of transposed convolution operations in these models \cite{durall2020watch}. This difference between natural and generated images facilitates the development of efficient and generator-agnostic detection tools \cite{grommelt2024fake}. However, the generalization of different detections remains a challenge in the relevant research in this field.
For instance, classifiers trained on GAN-based architectures often perform poorly when tested on VAE-based architectures, which is understandable since they employ different loss functions and image preprocessing methods \cite{wang2020cnn}.
\begin{figure}
\centering
	\includegraphics[width=\linewidth]{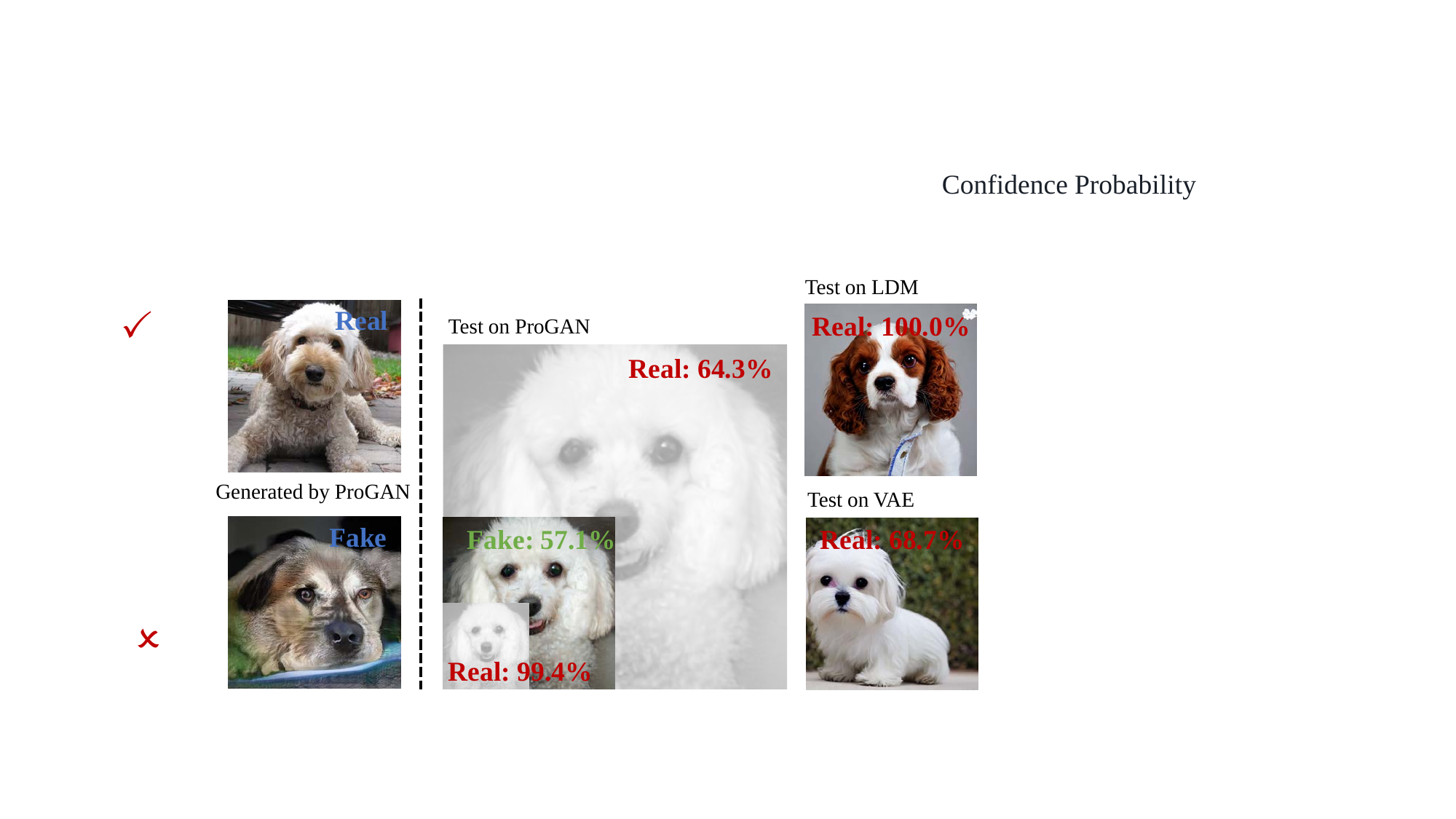}
	\caption{The generalization ability of generative image detection models is poor in multi-scale and cross-generator scenarios. In the figure, the left side represents the training dataset, where ProGAN is used as the generative model, while the right side shows the test dataset, along with the confidence scores for real or fake classifications.}
	\label{motivation}
\vspace{-0.2cm}
\end{figure}
The emergence of Denoising Diffusion Probabilistic Models (DDPMs)~\cite{ho2020denoising} has exacerbated this issue, which enabling the creation of highly realistic image editing. These models allow for easy manipulation of context-adaptive styles and lighting conditions \cite{nichol2021glide}. Models like DALL·E \cite{ramesh2022hierarchical} and Google Imagen \cite{saharia2022photorealistic} can even generate realistic videos directly from textual descriptions \cite{singer2022make}. Leveraging multimodal information, models under this paradigm are capable of generating smoother and more lifelike images. Moreover, these generative models exhibit a superior ability to approximate the frequency spectrum of natural images, resulting in a significant drop in the performance of many existing detectors on such images \cite{corvi2023intriguing, ojha2023towards}.

We find that a key factor affecting generalization is the model's lack of multi-scale understanding of objects. Image size variation refers to non-malicious degradations in real-world scenarios, such as re-compression \cite{grommelt2024fake} and resizing \cite{guillaro2023trufor}, commonly referred to as ``washing."  These operations can easily obscure traces of manipulation, making it difficult for detectors to identify forged content, as shown in Figure \ref{motivation}.
As can be seen, the model performs well only when the test images share the same resolution as the training images. However, scaling transformations, particularly downscaling, significantly degrade the model's performance, often leading to high-confidence misclassifications.
One of the main reasons for this phenomenon is that downscaling destroys critical discriminative features, and the model fails to learn sufficient generalizable features to ensure scale invariance. Furthermore, due to substantial differences in the generative processes of different models, the model also exhibits noticeable classification errors on data generated by unseen architectures.

% MSR leverages recent advancements in diffusion models, it generates multi-scale features for the target, incorporates a style-removal design to smoothly eliminate style information,
We aim to reach a fundamental conclusion: is the image an acceptable instance of reality or a product of generation? The motivation behind this objective is clear — if an image is known to be manipulated, we can either fully reject its authenticity or view it with partial skepticism. In this work, we propose a hierarchical retrospection refinement~(HRR) framework, which takes a significant step toward building a robust detector capable of reliably detecting generated images across generative paradigms while ensuring scale invariance. Specifically, it consists of two core designs, including Multi-scale Style Retrospection~(MSR) and Additive Feature Refinement~(AFR).
Inspired by the latest advancements in diffusion models, MSR generates multi-scale features for the target, incorporates a style-removal design to smoothly eliminate style information, and enhances the model's robustness and generalization across different generative paradigms and image scales through the introduction of pseudo-class augmentation.
AFR is designed based on the correntropy sparse additive machine, which is used to capture the intrinsic structures and patterns of the data, in which a sparse regularization term is employed to consistently reduce the impact of redundant features on predictive performance.
%Our approach achieves this goal through two core design elements, ultimately yielding a forgery detection framework with clear boundaries between real and fake categories and strong generalization capabilities.
Our main contributions are summarized as follows:
\begin{itemize}
\item We analyze the challenges in generated image detection tasks and identify that multi-scale information is a significant factor affecting model performance. We propose MSR to mitigate the learning bias introduced by specific datasets and generative models.
\item We designed AFR to capture the intrinsic structure and patterns of the data. By employing a sparse regularization term, it reduces the impact of redundant features on prediction performance, thereby enhancing the model's generalization ability.
\item Experiments on three benchmark datasets demonstrate that the proposed method shows clear advantages over baselines and achieving state-of-the-art results.
\end{itemize}

The rest of this paper is structured as follows. Section II makes and overview of related works relevant to our research. Section III presents a detailed description of our methodology. In Section IV, we conduct the experiments and analysis to validate our proposed method. Section V draws a conclusion.

\section{Related Work}
In this section, we provide an overview of the generalization research on generated image detection, review the progress of research on the relationship between image style and content, and explain the concept of additive models.

\subsection{Generalization Research of Generated Image Detection}

Research on generated image detection primarily focuses on the generalization across different generators. Current technologies \cite{guillaro2023trufor,chen2021image, liu2022pscc} are typically built on deep learning-based semantic segmentation frameworks, providing evidence of forgeries through local inconsistencies in color or mosaics~\cite{bammey2020adaptive}.  Using CLIP embeddings \cite{ojha2023towards} or inversion \cite{wang2023dire} has shown good performance on GAN-generated images. However, these methods do not generalized well to high-visual fidelity results generated by current diffusion models~\cite{huang2025pfb,rombach2022high}, even when images generated by specific diffusion models are included in their training data \cite{cazenavette2024fakeinversion}.
For this reason, DIRE~\cite{wang2023dire} was proposed, leveraging reconstruction error as a distinguishing feature for detecting diffusion-generated images. It is based on the assumption that, compared to real images, diffusion-generated images are more easily reconstructed by diffusion models. DIRE exhibits some generalizability to images generated by unseen diffusion models, showing good cross-model generalization capabilities. Inspired by DIRE, SeDID \cite{ma2023exposing} further utilizes the inherent distribution differences between natural images and diffusion-synthesized visuals for detecting diffusion-generated images. LaRE2 \cite{luo2024lare} further reveals that the loss from a single-step reconstruction is sufficient to reflect the differences between real and generated images.

However, most generators produce images with fixed sizes, which contrasts with the diverse size distribution observed in natural images. This discrepancy in size distribution may cause the detector to differentiate between natural and generated images based on size, significantly reducing the detector's robustness \cite{grommelt2024fake}. In this paper, we achieve scale-diverse generation through an ingenious design that introduces a scale perturbation mechanism during the generation process, dynamically adjusting the size distribution of generated images. This allows the generator to perceive and adapt to multi-scale variations during training, thereby enhancing detection performance for generated images.

% the presence of style may cause the model to be influenced by the training dataset, leading to the learning of dataset-associated features
\subsection{Style in Image}

Images consist of two main components: style and content. StyleDiffusion~\cite{wang2023stylediffusion} proposes a key insight that the definition of an image's style is significantly more complex than its content. Moreover,  style alone cannot reliably determine whether an image is generated or real. On the contrary, style may mislead the training of the model and affect the learning of key features of the dataset. As a result, the model may inherit the dataset's biases, disproportionately marginalizing certain groups \cite{cazenavette2024fakeinversion}.
To reduce the errors introduced by style, a constructive approach is disentangled representation learning, which aims to separate style from content. Gatys et al. explicitly defined high-level features extracted from a pre-trained Convolutional Neural Network (CNN) as content and feature correlations (i.e., Gram matrix) as style~\cite{gatys2016image}. This method achieved visually stunning results and inspired a large body of subsequent research~\cite{li2017universal,an2021artflow}. However, image disentanglement remains inexplicable and difficult to control~\cite{gatys2016image}. Other implicit content-style (C-S) disentanglement methods are often limited to the predefined domains of GANs (e.g., specific artistic styles~\cite{sanakoyeu2018style}), resulting in insufficient generalization and facing the same black-box issues related to interpretability and controllability~\cite{locatello2019challenging}.

Our HRR is based on the LDM, which can smoothly eliminate the style information from both content and style images, thereby reducing the bias introduced by style.

\subsection{Additive Models}

Additive Models (AMs) have attracted a great deal of attention due to the excellent performance on nonlinear approximation and the interpretability of their representation \cite{Meier2009High,yuan2023sparse}. The key characterization of AMs is the additive structure assumption of predictive functions.  The general form of an additive model is as follows:
\begin{equation}
	f(x)=\sum^{p}_{j=1}f_j(x_j),
\end{equation}
where $f(x)$ is the overall function, and $f_j(x_{j})$ is the component function with the $j$-th feature $x_j$.

Recently, to improve the interpretability of neural networks and the expressiveness and expansibility of additive models, many additive models have been proposed \cite{mueller2024gamformer,xu2023sparse,Chang22,enouen2022sparse,Agarwal21}.  Agarwal et al. proposed neural additive models (NAMs), which learned a neural network for each input feature \cite{Agarwal21}.  Furthermore,  to enable NAM to perform feature selection and improved the generalization ability, Xu et al. imposed a group sparsity regularization penalty (e.g., group Lasso) on the parameters of each sub-network, and proposed sparse neural additive models (SNAM) \cite{xu2023sparse}.  To narrow the gap in performance between the additive splines and the powerful deep neural networks, Enouen et al. \cite{enouen2022sparse}  proposed sparse interaction additive networks (SIAN), in which the necessary feature combinations can be efficiently identified by exploiting feature interaction detection techniques and genetic conditions. This allows training larger and more complex additive models.  Chang et al.  proposed Neural Generalized Additive Models (NODE-GAM) and Neural Generalized Additive Models plus Interactions (NODE-$\rm{GA^2M}$),  which improved the differentiability and scalability of additive models \cite{Chang22}. To eliminate the need for iterative learning and hyperparameter tuning , Mueller et al. proposed GAMformer,  which exploits the contextual learning to form shape functions in a single forward pass \cite{mueller2024gamformer}. However, those existing works still lack generalization ability when dealing with large-scale data, which limits the promotion of the models.

Our HRR is based on the additive models, which can accurately capture the intrinsic structure and patterns of the data, and reduce the impact of redundant features on prediction performance, thereby enhancing the model's generalization ability. In addition, our HRR is a preliminary exploration of additive models in large-scale data.

\section{Method} \label{method}
\begin{figure*}[hbpt]
	\begin{center}
		\includegraphics[width=0.99\textwidth]{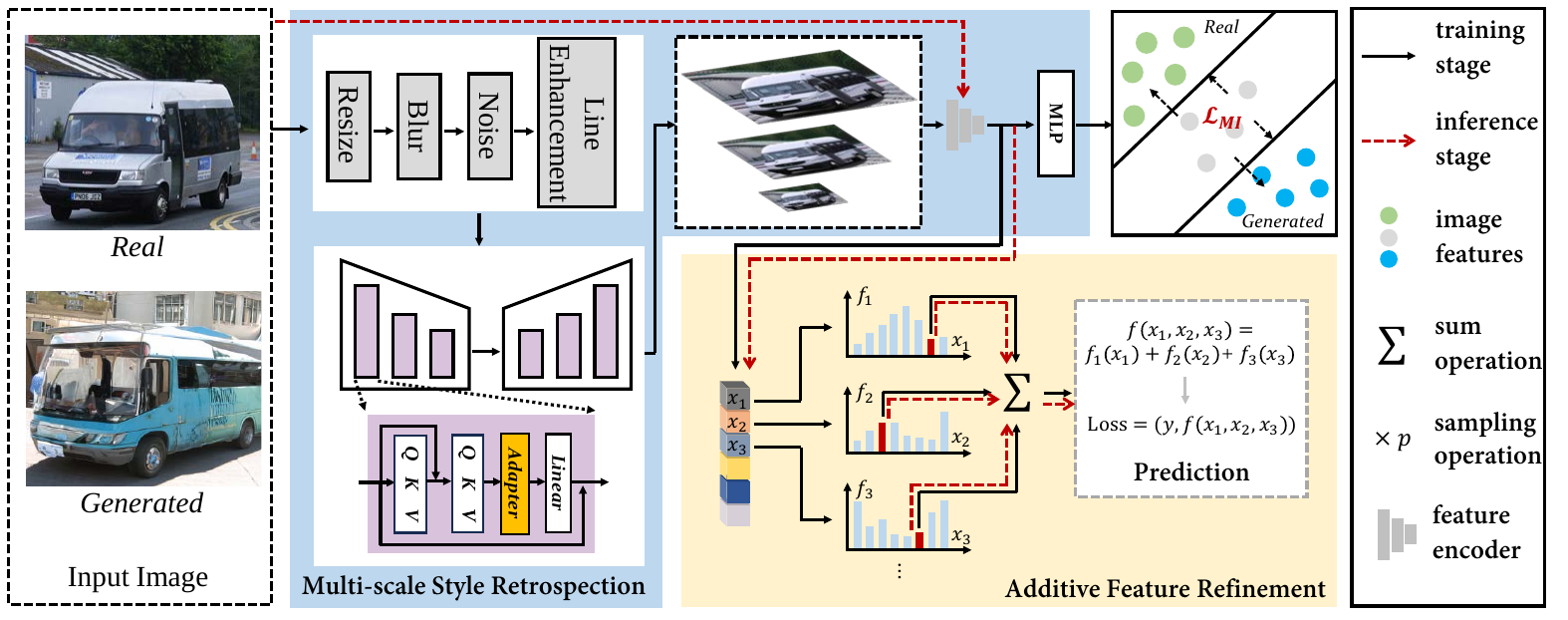}
	\end{center}
 \vspace{-0.2cm}
	\caption{Overview of HRR, it consists of two core modules: Multi-scale Style Retrospection~(MSR) and Additive Feature Refinement~(AFR).}
	\label{overview}
 \vspace{-0.4cm}
\end{figure*}

In this section, we first provide an overview of HRR, as shown in Figure~\ref{overview}, followed by a detailed explanation of how multi-scale feature extraction and style retrospection are implemented. Then, the implementation process of AFR based on Correntropy Sparse Additive Machine is thoroughly explained. Finally, we describe the overall optimization process.

\subsection{Multi-scale Style Retrospection~(MSR)}
The implementation of MSR involves two key steps: multi-scale generation and style retrospection.

To construct multi-scale features of the target, the most straightforward approach is to adopt a sampling-and-interpolation paradigm. However, recent studies \cite{wang2021real} have pointed out that directly employing global methods (e.g., modifying the degradation model or sharpening the entire ground truth) may introduce compression artifacts, as the network fails to effectively learn variations in line structures~\cite{wang2024apisr}.
To create a clear and coherent reference image after style normalization, we first resize the image and add noise to ensure edge continuity, minimizing the interference of jagged lines. 
Subsequently, line enhancement is applied. A Gaussian kernel-based sketch extraction algorithm is employed to extract edge maps from the sharpened GT~\cite{winnemoller2012xdog}, and an outlier filtering technique, coupled with a dilation method, is used to obtain images with clearer edges. Finally, edges are added to the image at the current scale, providing reliable contour constraints for subsequent diffusion generation.
The line enhancement can be formulated as follows:
\begin{equation}\label{Iline}
    I_{\text{Line-E}}  = h\big(g(f^n(I_{\text{GT}}))\big) + I_{\text{GT}}.
\end{equation}
% {\color{blue}
% \begin{align}
%     I_{\text{Sharp}}  & = f^n(I_{\text{GT}}), \tag{1} \\
%     I_{\text{Map}}    & = h\big(g(I_{\text{Sharp}})\big), \tag{2} \\
%     I_{\text{Line-E}} & = I_{\text{Map}} + I_{\text{GT}} . \tag{3}
% \end{align}
where $f^n$ is the sharpening function that recursively executes $n$ times, $g$ denotes edge detection, and $h$ stands for post-processing techniques of passive dilation with outlier filtering. %$\text{IMap}$ is a binary value map.

Then, LDM is employed to achieve multi-scale image generation and style retrospection.
The core of LDM is a conditional denoising network \( \epsilon_\theta \), which learns to reconstruct the latent representation from noisy data. The denoising network is optimized using the following loss function:
\begin{equation}
    \mathcal{L}_{\text{simple}} = \mathbb{E}_{z_0, \epsilon, t} \left[ \| \epsilon - \epsilon_\theta(z_t, t, c) \|^2 \right],
\end{equation}
where \( c \) represents the conditional input. In our model, input images at different resolutions are used as conditional information \( c \) provided to the denoising network.

The LDM restores images at different scales with clarity. Please note that during this process, we also use it to help eliminate domain-specific features of the input images and align them to a pre-trained domain \cite{kingma2018glow}. This is based on the assumption that images with different styles belong to distinct domains, but their content should share a common domain \cite{choi2021ilvr}. Therefore, LDM can be pre-trained on an alternative domain and then used to construct the content of the images. In this way, style features can ideally be removed, leaving only the content of the images.

Subsequently, this portion of the data is learned and embedded into both real and generated data through KL divergence \cite{kullback1997information}. Our goal is to increase the distance between these distributions during the learning process, so that the representations of generated images and real images in the feature space are as distinct as possible. We measured the similarity between the  distribution $P(x)$ of real image and  the distribution $Q(x)$ of generated image, which can be calculated as follows:
\begin{equation}\label{KL}
\mathcal L_{kl}= D(P \| Q) = \sum_{x} P(x) \log\left(\frac{P(x)}{Q(x)}\right).
\end{equation}
%Our goal is to minimize $\mathcal L_{kl}$ , which means that optimizing Eq.~(\ref{KL}) aims to make them as similar as possible. This knowledge transfer process allows valuable information learned from $\textbf{f}_{a}$ to be conveyed to $\textbf{f}^{x}$. 

\subsection{Additive Feature Refinement~(AFR)}
To accurately determine whether an image is an acceptable instance of reality or a product of generation, we propose AFR based on a sparse additive machine, which possesses dynamic feature selection capability.

%Let $\mathcal{Z}:=(\mathcal{X},\mathcal{Y})\subset\mathbb{R}^{p+1}$, where $\mathcal{X} \subset\mathbb{R}^{p}$ is a compact input space and $\mathcal{Y}=\{-1,1\}$ is the label set.

In AFR, we choose the basis-spline methods to estimate the component functions. The key concept is that the  component functions can be expressed as a linear combination of proper basis functions. 

%Let $\mathcal{H}= \mathcal{H}_1\oplus \mathcal{H}_2\oplus\cdots\oplus \mathcal{H}_p$ be the Hilbert space of functions that an additive form
%%%\begin{equation*}
%	f(x)=\sum^{p}_{j=1}f_j(x_{j}),
%\end{equation*}
%with $f_j\in \mathcal{H}_p,~j =1,\dots,p$

Denote the bounded and orthonormal basis functions on feature $x_j$ as $\{\psi_{jk}:k=1,\cdots,\infty\}$. Then, the component functions can be written as
%\begin{equation*}
$	f_j(x_{j})=\sum^{\infty}_{k=1}\alpha_{jk}\psi_{jk}(x_{j})$
%\end{equation*}
with coefficient $\alpha_{jk}, j=1,\cdots,p$. Actually, these basis functions is often truncated to finite dimension $d$. Then, we obtain
\begin{equation}\label{fj}
    f_j(x_{j})=\sum^{d}_{k=1}\alpha_{jk}\psi_{jk}(x_{j}).
\end{equation}
 
If the $j$-th variable is not truly informative, we expect that $\hat{\alpha}_{\mathbf{z},j}=(\hat{\alpha}_{\mathbf{z},j1},\ldots, \hat{\alpha}_{\mathbf{z},jd})^T\in \mathbb{R}^d$ satisfies $\|\hat{\alpha}_{\mathbf{z},j}\|_q=(\sum^d_{k=1}|\hat{\alpha}_{\mathbf{z},jk}|^q) ^{\frac{1}{q}}=0$. Inspired by this, we introduce the $\ell_{q,1}$-regularizer
\begin{small}
	\begin{eqnarray}\label{Omega}
		\mathcal L_q(f) = \inf\Big\{ \sum^{p}_{j=1} w_{j}\|\alpha_{j}\|_{q}:   f= \sum^{p}_{j=1}\sum^{d}_{k=1}\alpha_{jk}\psi_{jk}(x_{j}),\alpha_{jk}\in \mathbb{R}\Big\}
	\end{eqnarray}
\end{small}
as the penalty to address the sparsity of the output functions. 
Following the setting in \cite{WangY21, Chen17,yuan2023sparse},  the weight $w_{j}\equiv1,\;j=1,\cdots,p$ is used in AFR.  Such coefficient-based penalties have been widely used in support vector machines \cite{Huang2014Support,Wu2005SVM} with $q=1$ and sparse additive machines  \cite{Zhao12, WangY21, Chen17} with $q=2$.

Given $n$ training samples $\mathbf{z}=\{(x_i,y_i)\}^n_{i=1}$, the loss function of  AFR can be formulated as the following:
\begin{eqnarray}\label{fz}
    \mathcal L_{\sigma} +\lambda\mathcal L_q(f),
\end{eqnarray}
where $\lambda>0$ is a regularization parameter, and $\mathcal L_{\sigma}$  is the overall loss with the correntropy-induced loss (C-loss). The C-loss is defined as
\begin{equation}\label{Closs}
	\ell_{\sigma}(y,f(x))=\beta \Big[1-\exp\Big(-\frac{(1-yf(x))^{2}}{\sigma^{2}} \Big)\Big],
\end{equation}
where $\beta = (1-\exp(-1/\sigma^{2}))^{-1}>0$ and  $\ell_{\sigma}(y,0) =1$.

let $\Psi_{ji}=(\psi_{j1}(x_{ij}),\cdots,\psi_{jd}(x_{ij}))^{T}\in \mathbb{R} ^{d}$ and $\alpha_j = (\alpha_{j1},\cdots,\alpha_{jd})^{T}\in\mathbb{R}^{d}$. The objective function of AFR can be reformulated as
\begin{eqnarray}\label{alphaz}
    \min\limits_{\alpha_{j}\in\mathbb{R}^d,1\leq j\leq p}\frac{1}{n}\sum^{n}_{i=1}\ell_{\sigma}(y_i,  \sum^{p}_{j=1}\Psi_{ji}^T\alpha_{j})
    + \lambda \sum^{p}_{j=1} w_{j}\|\alpha_{j}\|_q.
\end{eqnarray}

 The objective function (\ref{alphaz}) can be converted into a weighted least squares problem via the half-quadratic (HQ) optimization \cite{Nikolova05}. Then, the ADMM strategy \cite{Boyd11} is employed to solve the transformed problem. Due to space limitation, a detailed optimization process is presented in Appendix \ref{opt}.

\subsection{Optimization}
In MSR, the loss function of the model is defined as:
\begin{eqnarray*}
 \mathcal L=\mathcal L_{ce}+ \gamma \mathcal L_{kl},
\end{eqnarray*}
It combines the standard cross-entropy loss with a KL loss as a regularization term, where $\gamma$ is the hyperparameter.

In AFR, the loss function can be defined as
\begin{eqnarray*}
 \mathcal L=\mathcal L_{\sigma}+ \lambda \mathcal L_q,
\end{eqnarray*}
where $\mathcal{L}_{\sigma}$ is the C-loss, and $\mathcal L_q(f)$  is  $\ell_{q,1}$-norm regularization term, where $\lambda$ is the hyperparameter.

\section{Experiments}
In this section, we separately evaluate the performance of HRR on three benchmark datasets. 
The datasets, implementation details, model configuration, and results are described below.
Ablations demonstrate the superiority of our method.

\subsection{Datasets and Experimental Setup}
Experiments are conducted on three generated image detection datasets.
%The details are summarized in Table~\ref{dataset}. 
% \begin{table}[htbp]\normalsize
% \caption{The splits of datasets. While \textit{C}$_{all}$ is the number of total subclasses, \textit{C}$_{train}$, \textit{C}$_{test}$ represent the number of training and testing images in the dataset respectively.}
% 
% \centering
%  \begin{tabular}{l|c|c|c|c}
% \hline
%  \textbf{Dataset}   & \textit{C}$_{all}$  & \textit{C}$_{train}$  & \textit{C}$_{test}$ & Image size\\  
% \hline
% DIRE ~\cite{wang2023dire}		  & 23929   & 24633  & 555   & 256$\times$256\\
% ForenSynths~\cite{wang2020cnn}    & 720000  & 4000   & 200   & 256$\times$256\\ 
% cocoFake\cite{amoroso2023parents} & 197259  & 24567  & 24567 & 512$\times$512\\  
% \bottomrule
% 	\end{tabular}
% 	\label{dataset}
% \end{table}
\textbf{DIRE}\cite{wang2023dire} contains real images from LSUN-Bedroom and ImageNet, as well as images generated by corresponding pretrained diffusion models. According to the type of diffusion models, the images are categorized into three classes: unconditional, conditional, and text-to-image. The experiments in this paper are based on the unconditional generation results, with the training data involving four generators: DDPM, iDDPM, ADM, and PNDM.
The number of training and testing images in each subset constructed under each generator is 42,000.

\textbf{ForenSynths}\cite{wang2020cnn} includes results generated by 11 different generators or techniques, covering three unconditional GANs (Pro/Style/Big GAN), three conditional GANs (Gau/Cycle/Star GAN), as well as generation results for low-level vision tasks. For each type of result, the number of real and fake images is approximately balanced. The training set is constructed using ProGAN. This dataset is constructed using 20 models, each trained on a different object category, generating 36,000 training images and 200 validation images. For each model, the training and validation sets contain an equal number of real and synthesized images. In total, the dataset consists of 720,000 training images and 4,000 validation images.

\textbf{cocoFake} dataset consists of two parts, including cocoFake\cite{amoroso2023parents} and COCO\cite{lin2014microsoft} dataset. Specifically, cocoFake includes 414,113 training samples and 202,654 validation samples, generated by Stable Diffusion v1.4 using textual prompts derived from the COCO dataset. Compared to existing datasets for deep fake detection, cocoFake exhibits greater diversity, uniform coverage of semantic classes, and can be easily scaled to a larger size. On the other hand, the COCO dataset contains 82,783 training samples and 40,504 validation samples. We combine the entire COCO dataset with 123,287 selected samples from cocoFake to form a new training set, which is then split into training, validation, and testing subsets following an 8:1:1 ratio.

\subsection{Implementation Details}
\textbf{Training Setting.} In the image generation process, we adopted the SD-XL-refiner-1.0 model~\cite{podell2023sdxl}. The input images are real images totaling 80k sampled from DIRE dataset with prompt left blank and input images resized to several size. To build a generated image classifier, we used Adam optimizer with 1e-4 learning rate. For all datasets, batchsize = 256 and the same early stopping strategy takes effect. After we trained each model, we passed all the images within the dataset through the model to get the feature extracted by the backbone which is then passed to the AFR to get refined for classification. The code is based on PyTorch and was run on one NVIDIA RTX 3090 GPU.

\textbf{Metrics.} To evaluate the performance of the proposed model for  generated image detection, we utilize two metrics: accuracy (Acc.) and average precision (AP). Acc. measures the classification accuracy of all classes, while AP represents the average precision across all recall, specifically the area under the PR curve (AUC). Higher Acc. and AP values indicate superior performance. We use both metrics to get a comprehensive evaluation of the model.

\subsection{Model Configuration}
Model configuration experiments is conducted on the ForenSynths dataset to verify the validity of the individual component and to determine the hyperparameters. For intuitive visualization, all results are normalized to the range of 0 to 1.

\begin{figure}[hbpt]
	\centering
	\begin{tabular}{cc}
		\subfigure[KL Loss ($\gamma$)]{\includegraphics[width=0.5\linewidth]{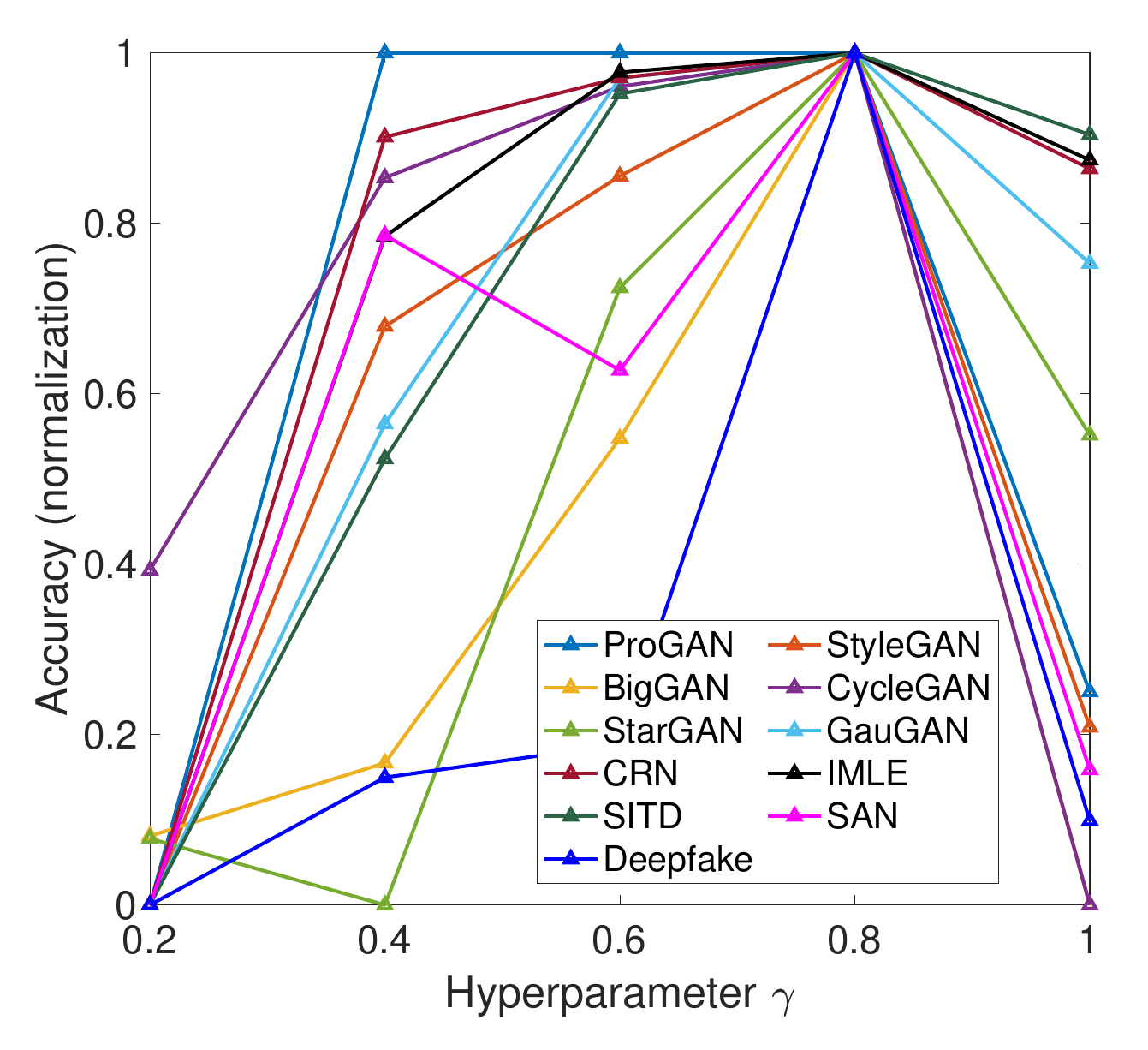}}\hspace{-4mm}
		&		
		\subfigure[$\ell_{q,1}$-norm Regularizer ($\lambda$)]{\includegraphics[width=0.5 \linewidth]{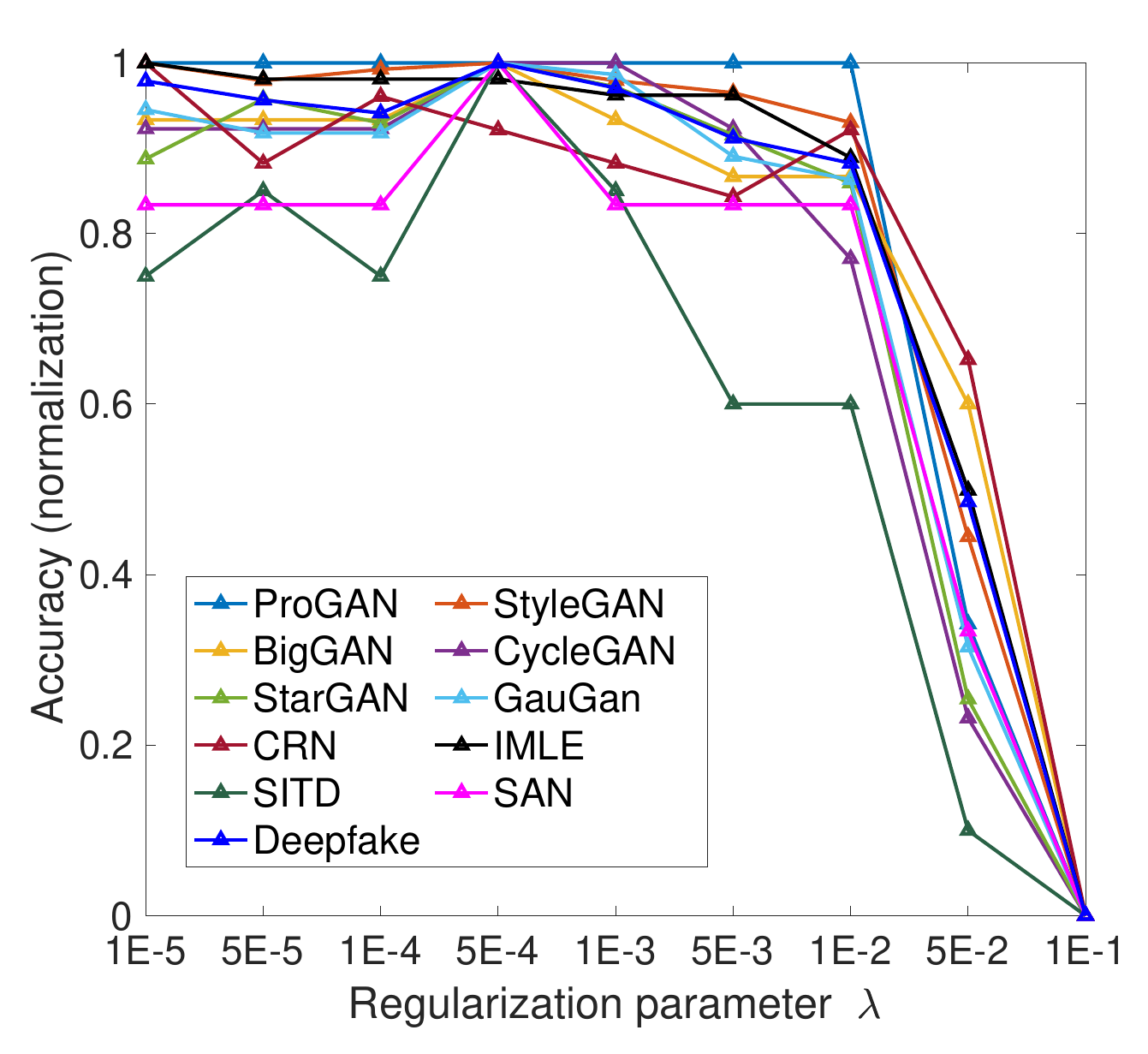}}\hspace{-3mm}\\
	\end{tabular}  
     \caption{Evaluation of the hyper-parameter $\gamma$ of KL-Loss and the regularization parameter $\lambda$. The horizontal axis represents the value of the hyperparameter $\gamma$  or $\lambda$, while the vertical axis indicates model accuracy.}
\vspace{-0.2cm}
\label{parameter}
\end{figure}

Figure~\ref{parameter}(a) shows that KL Loss provides effective auxiliary support for MSR, we investigate the impact of varying the proportion of KL loss in the loss function. As can be seen, when $\gamma$ is small, the contribution of KL loss is minimal compared to the classification loss, and its auxiliary effect is not significant. As the proportion of KL loss increases, the performance of MSR gradually improves. However, if $\gamma$ becomes too large, the center of gravity of MSR shifts, resulting in a decline in performance. Therefore, in this work, we set this parameter to 0.8.
Figure~\ref{parameter}(b) shows the effect of different value of $\lambda$ in AFR module.
It can be seen that  the value 5E-4 are the best for model learning. In particular,  when the lambda is larger than 1E-2, the prediction performance drops sharply. One important reason considering that a smaller $\lambda$ makes the regularization less penalize the complexity of the model, and the model can capture more features of the data. As the $\lambda$ increases, the regularization penalizes the model complexity more, which forces the model weight (e.g., the parameter $\alpha$) to be smaller. This may lead to underfitting of the model and performance degradation.

\begin{table*}[htbp]\normalsize
\caption{
Cross-generator generalization results. We show the Acc. (\%) and mAP (\%) of various classifiers from baseline and ours, tested across 8 generators. The training set used in the experiment is LSUN-B.
%The highest value is highlighted in black.
%The best and the second best results are marked in {\color[HTML]{C00000}red} and {\color[HTML]{0070C0} blue}, respectively. The symbol '-' indicates no results.
}
 
\centering
\begin{tabular}{p{1.6cm}|p{1.7cm}<{\centering}|p{0.7cm}<{\centering}p{0.8cm}<{\centering} p{1cm}<{\centering}p{0.8cm}<{\centering}p{0.7cm}<{\centering} p{0.7cm}<{\centering}  p{0.7cm}<{\centering} p{1.7cm}<{\centering} p{0.7cm}<{\centering}}%{c|c|ccccccccc}
\hline
 \textbf{\multirow{2}{*}{Method}}  & \textbf{\multirow{2}{*}{Gen. Model}}
 & \multicolumn{8}{c}{\textbf{Testing Individual Diffusion Generators}} & \multirow{2}{*}{mAP}\\ \cline{3-10}
&  &ADM	&DDPM  &iDDPM	&PNDM	&sdv1	&sdv2	&LDM &VQDiffusion &\\  
\hline
\multirow{4}{*}{Efficient-net}
     & ADM      &95.65	&92.57  &98.70  &98.20  &44.67  &76.95  &85.9	&88.00&93.91\\
     & iDDPM    &\bf 94.93	&89.84	&97.80	&97.10	&\bf 44.22	&\bf 79.95	&\bf 82.7	&85.90&92.10\\
     & PNDM     & 8.31	&70.57	&76.20	&99.40	& 6.50	& 0.15	& 7.7	&50.00&84.30\\
     & StyleGAN & 2.05	&10.54	& 8.70	&33.50	& 4.16	& 0.05	& 9.6	&40.80&78.43\\
                         \hline
\multirow{4}{*}{CNNDet}
     & ADM      &99.55   &93.09  &\bf 99.40	&99.60	&55.68  &97.80	&\bf 99.20	&\bf 99.00 &92.65\\
     & iDDPM    &92.10	&\bf 97.14  &\bf 99.90	&\bf 99.50	&38.92	&78.70	&77.35	&87.65&84.56\\
     & PNDM     &55.05	&60.76  &71.95	&\bf 99.95	&21.72	&60.55	&51.35	&57.70&86.07\\
     & StyleGAN &50.10	&36.45  &50.25	&53.10	&38.12	&68.50	&52.50	&62.60&66.27\\
\hline
% \multirow{4}{*}{Swim-Vit}
%      & ADM      &96.87   &99.22  &99.50  &99.90  &70.94  &99.90  &100.0  &100.0&\\
%      & iDDPM    &17.77 	&99.61 	&99.90 	&100.0 	& 5.99 	&24.65 	&90.00 	&100.0&\\
%      & PNDM     & 6.98 	&64.97 	&34.10 	&100.0 	& 9.13 	&24.75 	&97.00 	&100.0&\\
%      & StyleGAN &16.88 	&92.06 	&82.20 	&99.60 	&30.86 	&48.70 	&99.10 	&100.0&\\
%                          \hline
% \multirow{4}{*}{FakeInversion}
%      & ADM      &- &- &- &- &- &- &- &-&\\
%      & iDDPM    &- &- &- &- &- &- &- &-&\\
%      & PNDM     &- &- &- &- &- &- &- &-&\\
%      & StyleGAN &- &- &- &- &- &- &- &-&\\
%                          \hline                         
\multirow{4}{*}{HRR(Ours)}
    &ADM &\bf 99.15	&\bf 93.33	&99.20	&\bf 99.70	&\bf 68.51	&\bf 98.15	&98.55	&97.10	& \bf 96.66 \\
    %&ADM &97.60	&90.38	&98.15	&98.55	&64.59	&96.15	&97.20	&96.95&\\
    &iDDPM	&93.65	&96.44	&99.50	&98.60	&42.16	&76.00	&81.55	&\bf 93.95  &\bf 97.51\\
    &PNDM	&\bf 56.30	&\bf 72.74	&\bf 84.45	&99.80	&\bf 59.62	&\bf 73.80	&\bf 65.25	&\bf 81.80  &\bf 91.33\\
    &StyleGAN  &\bf 54.55	&\bf 62.84	&\bf 58.95	&\bf 64.60	&\bf 48.78	&\bf 73.95	&\bf 61.20	&\bf 81.85  &\bf 82.89\\
\hline
	\end{tabular}
	\label{DIRE}
\end{table*}

\subsection{State-of-the-art~(SOTA) Comparison}
In this section, we compare our HRR with several classic approaches, include Efficient-net~\cite{tan2019efficientnet}, Swim-Vit~\cite{liu2021swin}, and CNNDet~\cite{wang2020cnn}. The best results are highlighted in bold. % while the second-best results are underlined. , and FakeInversion~\cite{cazenavette2024fakeinversion}

\begin{table*}[htbp]
\caption{
Cross-generator generalization results. The generation model is ProGAN. We show the Acc. (\%) and mAP (\%) of various classifiers from baseline and ours, tested across 10 generators.  %The highest value is highlighted in black.
%The best and the second best results are marked in {\color[HTML]{C00000}red} and {\color[HTML]{0070C0} blue}, respectively. The symbol '-' indicates no results.
}\centering
    \begin{tabular}{p{1.7cm}<{\centering}|p{0.8cm}<{\centering}p{0.9cm}<{\centering}p{0.8cm}<{\centering} p{1cm}<{\centering}p{0.8cm}<{\centering}p{0.8cm}<{\centering} p{0.8cm}<{\centering}  p{1cm}<{\centering} p{0.8cm}<{\centering} p{1.4cm}<{\centering}p{0.7cm}<{\centering}}
    %{c|ccccccccccc}
    \hline
         \textbf{\multirow{3}{*}{Method}}   
         & \multicolumn{10}{c}{\textbf{Individual Test Generators}}  & {\multirow{3}{*}{mAP}}\\ \cline{2-11} 
      %   &\shortstack{Pro-\\GAN} &\shortstack{Style-\\GAN} &\shortstack{Big-\\GAN}	&\shortstack{Cycle-\\GAN}	&\shortstack{Star-\\GAN}	&\shortstack{Gau-\\GAN}	&CRN	&IMLE		&SAN	&Deepfake\\         
      &\makecell[c]{Pro-\\GAN}  &\makecell[c]{Style-\\GAN} &\makecell[c]{Big-\\GAN}	&\makecell[c]{Cycle-\\GAN}	&\makecell[c]{Star-\\GAN}	&\makecell[c]{Gau-\\GAN}	&\makecell[c]{CRN}	&\makecell[c]{IMLE}	&\makecell[c]{SAN}	&\makecell[c]{Deepfake}&\\ %&\makecell[c]{SITD}	
        \hline
        \multirow{1}{*}{Efficient-net} &99.70	&68.58  &57.20	&79.67  &72.33	&81.40	&72.40	&84.22	  &50.22	&50.93	&73.38\\ %&73.05
        
        \multirow{1}{*}{CNNDet}  &99.92	&72.85	&60.29	&84.29	&85.47	&80.81	&87.51	&95.01		&50.46	&52.85 & 74.09 \\ %&78.33        
        %\multirow{1}{*}{Swim-Vit}& \multirow{1}{*}{ProGAN}&100.00	&86.38	&91.68	&98.94	&97.20	&96.66	&97.59	&98.04	&66.94	&53.88	&60.07	\\
        %\multirow{1}{*}{FakeInversion}& \multirow{1}{*}{ProGAN}  &-	&-	&-	&-	&-	&-	&-	&-	&-	&- &\\        
        \multirow{1}{*}{HRR(Ours)} &\bf 99.92  &\bf 85.36  &\bf 63.20  &\bf 86.26   &\bf 86.22	&\bf 84.53	&\bf 99.26	&\bf 98.40	&\bf 55.25	&\bf 56.30 & \bf 75.39\\ % &\bf 91.94	
        \hline
    \end{tabular}\label{ForenSynths}
\end{table*}

\subsubsection{Experimental results on DIRE dataset}
The comparison results are presented in Table~\ref{DIRE}. 
It can be observed that in cross-generator experiments, the model exhibits a significant advantage on test subsets generated by the same generator as the training set. Taking the ADM generator as an example, both EfficientNet and CNNDet achieve the highest accuracy on the ADM subset. Compared to models trained with the second-best generator, they obtain improvements of 0.72\% and 7.45\%, respectively.
Notably, this is not an isolated case, as similar results are observed with the iDDPM and PNDM generators. Although PNDM achieves relatively high accuracy in certain cross-generator evaluations, it still attains the best performance on its own subset. This demonstrate the widespread challenge of generalization difficulty.
In contrast, our model demonstrates stable performance in cross-generator evaluations, with particularly significant improvements when trained on images generated by StyleGAN and tested on images generated by diffusion models. Specifically, compared to CNNDet, our model achieves accuracy improvements of 4.45\%, 26.39\%, 8.7\%, 11.5\%, 10.66\%, 4.45\%, 5.45\%, 8.7\%, and 19.25\%, respectively.

\subsubsection{Experimental results on ForenSynths dataset}
The comparison results are presented in Table~\ref{ForenSynths}. 
The test dataset of ForenSynths consists of multiple subsets, each corresponding to images generated by different generators (GAN or CNN variants). Among the GAN-based test subsets, BigGAN exhibits the lowest performance, partly because this subset is primarily designed for high-resolution image generation. As previously discussed, cross-generator detection models are highly sensitive to scale variations, which can easily lead to misclassification. However, benefiting from our multi-scale training strategy, the model achieves the best performance on the BigGAN test set, outperforming the second-best model CNNDet by 2.91\% in accuracy. 
The ForenSynths dataset also includes various generation methods and models for image synthesis. Our algorithm achieves a significant improvement on the CRN subset, outperforming CNNDet by 11.75\%. CRN is a generative model that combines convolutional and recurrent neural networks to generate images with temporal or sequential dependencies.
This improvement highlights the effectiveness of our approach in processing high-dimensional temporal or sequential  data. SAN is a generative network that incorporates a self-attention mechanism, enabling it to produce images with higher detail and coherence. It is used to evaluate the capability of image detection models in capturing global contextual relationships. Despite the high difficulty of this subset, our model still achieves the best performance, outperforming the previous approach by 4.79\%.

\subsubsection{Experimental results on cocoFake dataset.}
The comparison results are presented in Table~\ref{cocoFake}. 

\begin{table}[htbp]\small
\caption{Comparison results. Acc. (\%) and AP (\%) (Acc./AP in the table) are presented. 
%The highest value is highlighted in black.
}\centering
    \begin{tabular}{c|c|c}
        \hline
        \textbf{\multirow{2}{*}{Method}}   & \textbf{\multirow{2}{*}{Gen. Model}} & \multicolumn{1}{c}{\textbf{Test Generator}} \\ \cline{3-3}
        &  &LDM	\\     \hline
        Efficient-net  & LDM &  95.93~/~99.64 \\
        CNNDet         & LDM &  91.31~/~98.31 \\
        %FakeInversion  & LDM &  - ~/~ - \\
        HRR(Ours)      & LDM &  \textbf{96.53}~/~\textbf{99.94} \\
        \hline
    \end{tabular}\label{cocoFake}
\end{table}
It can be observed that, since the COCOFake dataset does not involve the challenge of model generalization across different generators, training a binary classification model can effectively achieve the detection objective. However, our method consistently achieves the best performance. Compared to SOTA approaches, it improves accuracy by 0.6\% while maintaining a competitive advantage in terms of AP. %{\color{red} Notably, FakeInversion is a specialized model designed specifically for detecting images generated by Stable Diffusion. Our superior performance demonstrates that HRR exhibits better generalization capability.}

\begin{table*}[htbp]\small
\caption{Ablation studies of the HRR on ForenSynths dataset. The generation model is ProGAN.} 
\centering
\begin{tabular}{c|ccccccccccc}
    \hline  
        \textbf{\multirow{3}{*}{Method}}     & \multicolumn{10}{c}{\textbf{Individual Test Generators}}  & {\multirow{3}{*}{mAP}}\\ \cline{2-11} 
           &\makecell[c]{Pro-\\GAN}  &\makecell[c]{Style-\\GAN} &\makecell[c]{Big-\\GAN}	&\makecell[c]{Cycle-\\GAN}	&\makecell[c]{Star-\\GAN}	&\makecell[c]{Gau-\\GAN}	&\makecell[c]{CRN}	&\makecell[c]{IMLE}	&\makecell[c]{SAN}	&\makecell[c]{Deepfake}\\ \hline 
     %   &\shortstack{Pro-\\GAN} &\shortstack{Style-\\GAN} &\shortstack{Big-\\GAN}	&\shortstack{Cycle-\\GAN}	&\shortstack{Star-\\GAN}	&\shortstack{Gau-\\GAN}	&CRN	&IMLE		&SAN	&Deepfake\\         
        \multirow{1}{*}{Baseline}&99.92  &72.85  &60.29  &84.29	&85.47	&80.81	&87.51	&95.01	&50.46	&52.85 &74.09\\ %&78.33	        
        \multirow{1}{*}{w/o AFR} & \bf 99.97  &78.37  &62.98  &80.24   &82.51   &\bf 88.32  &94.61  &95.11   &51.22  &53.39 & 74.76 \\ %&87.77         
        \multirow{1}{*}{w/o MSR} &99.96  &85.08  &62.03  &85.83   &85.49   &84.31	&96.48	&97.39	&53.34	&53.52 &75.03\\ %&78.33	
        \multirow{1}{*}{HRR}   &99.92  &\bf 85.36  &\bf 63.20  &\bf 86.26   &\bf 86.22	&84.53	&\bf 99.26	&\bf 98.40		&\bf 55.25	&\bf 56.30 & \textbf{75.39}\\  \hline
    \end{tabular}\label{Ablation}
\end{table*}

\begin{table*}[htbp]\small
\caption{Multi-scale performance analysis of the HRR on ForenSynths dataset. 
All test data is not operated with crop.}

\centering
    \begin{tabular}
    {p{1.3cm}<{\centering}|p{1.3cm}<{\centering}|p{0.7cm}<{\centering}p{0.8cm}<{\centering}p{0.7cm}<{\centering} p{0.9cm}<{\centering}p{0.7cm}<{\centering}p{0.7cm}<{\centering} p{0.7cm}<{\centering}  p{0.8cm}<{\centering} p{0.7cm}<{\centering} p{1.3cm}<{\centering}p{0.6cm}<{\centering}}
    %{c|ccccccccccc}
\hline
 \textbf{\multirow{3}{*}{Method}}   & \textbf{\multirow{3}{*}{Size}}
 & \multicolumn{10}{c}{\textbf{Individual Test Generators}}  & {\multirow{3}{*}{mAP}}\\ \cline{3-12} 
     &    &\makecell[c]{Pro-\\GAN}  &\makecell[c]{Style-\\GAN} &\makecell[c]{Big-\\GAN}	&\makecell[c]{Cycle-\\GAN}	&\makecell[c]{Star-\\GAN}	&\makecell[c]{Gau-\\GAN}	&\makecell[c]{CRN}	&\makecell[c]{IMLE}	&\makecell[c]{SAN}	&\makecell[c]{Deepfake}\\ \hline  
%&  &\shortstack{Pro-\\GAN} &\shortstack{Style-\\GAN} &\shortstack{Big-\\GAN}	&\shortstack{Cycle-\\GAN}	&\shortstack{Star-\\GAN}	&\shortstack{Gau-\\GAN}	&CRN	&IMLE		&SAN	&Deepfake\\ 
\multirow{4}{*}{CNNDet}
& $512\times 512$ &99.97 &89.31 &86.82  &97.19  &95.77  &89.76  &99.78  &99.77  &50.22  &50.41 &93.62\\ %&63.61 
& $256\times 256$ &99.98 &78.92 &59.85  &84.44  &85.27  &82.51  &97.85  &98.02  &50.01  &50.12 &78.69\\ %&91.67 
& $128\times 128$ &96.79 &60.99	&53.91	&69.53	&67.53	&75.89	&79.01	&78.53	&50.01	&51.11 &68.33\\ %&80.28
& $64 \times 64$  &80.81 &55.56	&52.95	&59.23	&58.18	&69.31	&65.36	&66.26	&49.77	&50.78 &60.82\\ %&74.44	
\hline
%\multirow{4}{*}{Swim-Vit}& \multirow{4}{*}{ProGAN}&100.00	&86.38	&91.68	&98.94	&97.20	&96.66	&97.59	&98.04	&66.94	&53.88	&60.07	\\
%                    &                             &100.00	&87.62	&84.25	&97.35	&97.52	&93.94	&99.22	&99.55	&67.78	&50.91	&52.40	\\
%                    &                             &99.92	&85.66	&68.88	&89.93	&92.95	&80.94	&98.21	&98.53	&67.78	&50.46	&51.64	\\
%                    &                             &99.61	&86.79	&68.90	&83.20	&90.05	&83.37	&95.43	&96.55	&66.11	&50.00	&51.36	\\

%\multirow{1}{*}{HRR(ours)}& \multirow{1}{*}{ProGAN}&99.91  &85.36  &63.20  &86.26   &86.22	&84.53	&99.26	&98.40	&91.94	&55.25	&56.30 \\
\multirow{4}{*}{HRR(ours)}
&$512\times 512$ &99.93	&88.87	&89.45	&97.24	&91.40	&94.45	&99.28	&98.74	&57.76	&51.47&96.99
\\
&$256\times 256$ &99.90	&78.75	&63.18	&86.22	&86.22	&84.52	&99.22	&99.35	&51.14	&55.62&93.08\\
&$128\times 128$&96.51	&62.69	&55.60	&71.46	&71.81	&75.31	&83.23	&89.68	&50.46	&51.19 & 91.44\\
&$64 \times 64$&77.04	&55.25	&55.40	&60.67	&58.95	&69.36	&66.55	&75.58	&50.00	&51.42 & 82.21\\
\hline
	\end{tabular}
	\label{size_compare}
\end{table*}

\subsection{Ablation Study}
To evaluate the proposed method, an ablation study was conducted on ForenSynths dataset.
Firstly, the performance of the feature extraction model was evaluated as the baseline.
``w/o AFR" means that only MSR is used to generate multi-scale style-normalized features for learning. ``w/o MSR" means that we only process the final extracted features of the model, without providing scale information, allowing the model to learn fully. The complete model is presented using HRR. The experimental results were presented in Table~\ref{Ablation}.

It can be observed that, compared to the baseline results, supplementing feature learning with multi-scale, style-consistent image data generated by MSR effectively enhances the model's understanding of instance contours and structural information. This approach achieves an average performance improvement of 8\% on datasets generated by StyleGAN, GauGAN, CRN, and SITD, while also yielding favorable results on data generated by other methods.
Next, we evaluated the effectiveness of the proposed AFR module. The results show that, built upon the baseline, this module consistently improves accuracy, demonstrating its ability to focus on learning more discriminative features of the target. When both modules are used together, the model achieves its best performance.

\subsection{Complexity and Generalization Analysis}
For simplicity, we first focus on the analysis of the computational cost for the proposed method in the $t$th iteration. According to the optimization in (\ref{objB}), the computational complexity of $b$ is $\mathcal{O}(n)$. According to our optimization for  (\ref{alpha_t1}), the main computational cost is from the inverse of $\Phi^T{\rm diag}(-b)\Phi$. Hence, the computational complexity of $\alpha$ is close to $\mathcal{O}(dpn^2+d^2p^2n+d^3p^3)$. In addition, we can easily derive the computational complexity of
$\vartheta$ and $\mu$ from  (\ref{betaS}) to  (\ref{mu}), where the computational complexity of $\vartheta$  and $\mu$ are both approximately  $\mathcal{O}(dp)$. Therefore, the computational complexity of this ADMM in the $t$th iteration is close to $\mathcal{O}(dpn^2+d^2p^2n+d^3p^3)$. Hence, the computational complexity of this ADMM  is  about $\mathcal{O}(T(dpn^2+d^2p^2n+d^3p^3))$, where $T$ is the iterations of ADMM procedure. To summarize, the computational complexity of proposed method  is  about $\mathcal{O}(T(T(dpn^2+d^2p^2n+d^3p^3))+n)$, where $T$ is the iterations of the outer optimization. 

We further analyze the generalization capability of the model. Specifically, while keeping the training dataset unchanged, we apply different scale transformations to the test images to evaluate the model's performance in handling multi-scale variations. This approach allows us to assess the model's robustness and adaptability to changes in image scale.
The comparison results are presented in Table~\ref{size_compare}. 

It can be observed that when the test image size matches the training image size, the model maintains a relatively high level of performance. However, both CNNDet and Efficient-net struggle to handle the challenge of multi-scale variations, leading to significant performance degradation. A key factor contributing to this decline, as discussed earlier, is that the whitening operation on images can easily obscure modification traces, making it difficult for the model to make correct judgments.
In contrast, benefiting from our MSR design, the model is able to comprehensively learn the discriminative features of the target, mitigating the interference caused by scale variations. As a result, it achieves the best performance.

\subsection{Visualization}
\begin{figure*}[htbp]
	\centering
	\begin{tabular}{c}
		\subfigure[CAM visualization on ForenSynths dataset.]{\includegraphics[width=0.97 \linewidth]{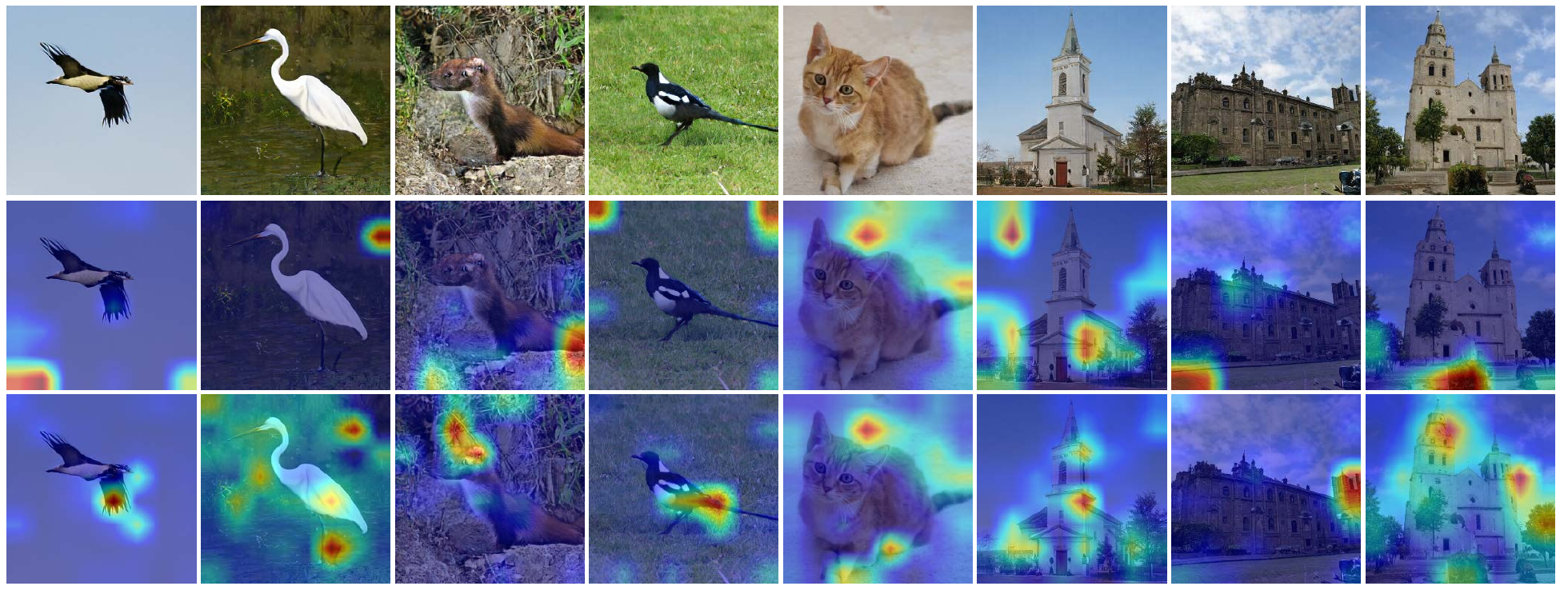}}\hspace{-0mm} \vspace{-0.2cm}
		\\	    
		\subfigure[CAM visualization on DIRE and cocoFake datasets.]{\includegraphics[width=0.97 \linewidth]{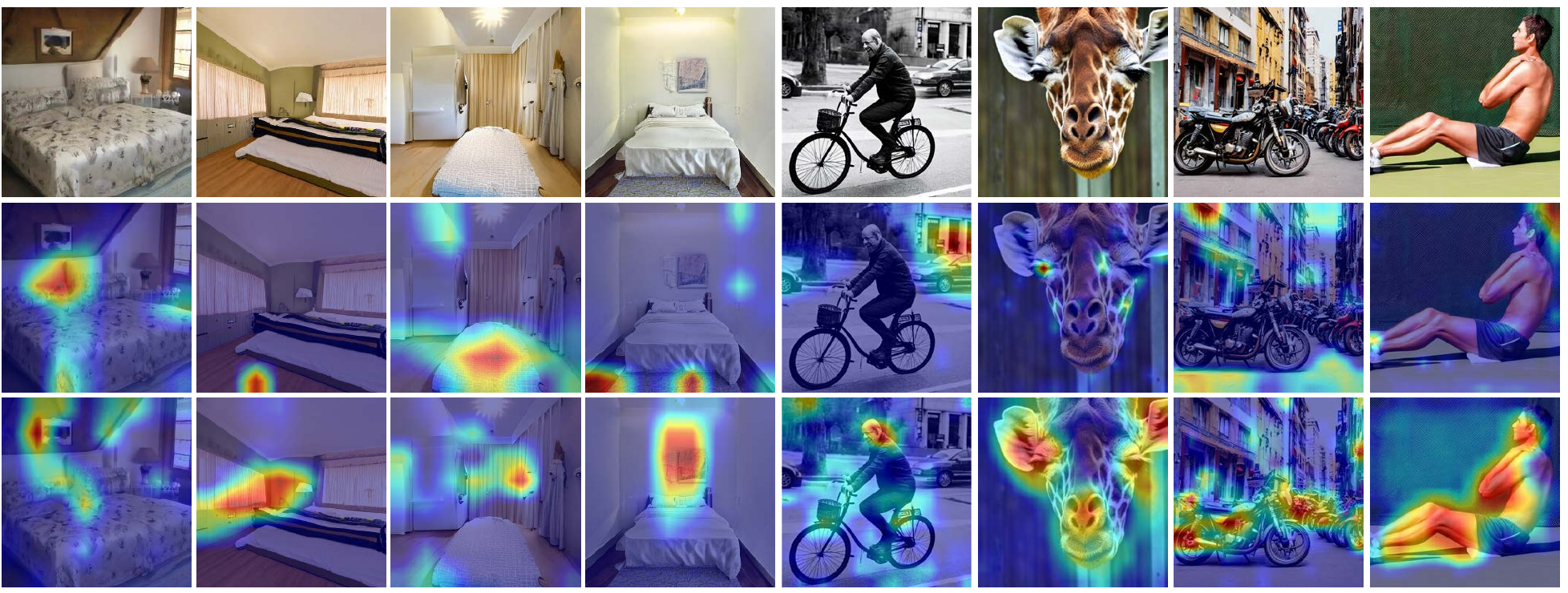}}\hspace{-0mm}
	\end{tabular}
\vspace{-0.2cm}
    \caption{CAM visualization. In each subfigure, the first row represents the original image, and the second to third rows represent the results of the baseline and our method, respectively.}
    \label{CAM}
\end{figure*}

We qualitatively demonstrate the impact of the proposed algorithm on model learning in generated image through CAM~\cite{zhou2016learning} visualization. The results are presented in Figure~\ref{CAM}.
Several interesting phenomena can be observed from Figure~\ref{CAM} (a). Images generated by GAN-based models tend to exhibit repetitive structures in local regions, such as the eyes of otters, cat paws, or the abrupt appearance of half floors or spires in buildings. This may partially explain the periodic and grid-like patterns observed in their frequency spectra, since the periodicity in the frequency domain reflects the regularity in spatial structures within the image, which further reinforces the artificial nature of the generated content.
When dealing with this type of generated image, the baseline struggles to uncover the underlying issues solely from the image domain. For example, baseline models may not be able to recognize the repetitive artifacts of specific object parts because the repetition might not always be overtly visible in the image's global context. Additionally, when these models do attempt to locate anomalies, their detection is often scattered—localized in random regions that do not provide a coherent understanding of the issue.  This scattered localization suggests the model is confused or uncertain about the true nature of the problem, and its decisions lack consistency.
In contrast, our HRR enables multi-scale learning based on CNN contour information and evaluates features across dimensions, efficiently rejecting unreasonable regions, such as inconsistent wing-flapping poses in flying birds, discontinuous legs in waterfowl, repeated eyes in otters, and repetitive architectural structures.

Figure~\ref{CAM} (b) presents the results generated by a diffusion model. With the aid of multi-modal information, models in this paradigm are capable of generating images with smoother and more realistic appearances. Therefore, generating images is recognized as a relatively challenging task. One of the reasons for the success of our method lies in its sensitivity to contours within the image, such as the repeated wrinkles and curved walls shown in the figure. Furthermore, some generators (e.g., MidJourney) produce images with certain characteristic patterns, such as color combinations and compositional styles, forming a distinct visual style. Although the details are well-handled, thanks to our style retrospection design, the model can detect generation traces and thereby make accurate distinctions.

\section{Conclusion}
Understanding the differences between real and generated images from multi-scale, style-agnostic representations, constructing a compact expression of the target, and capturing the intrinsic structure and patterns of the data are key steps toward achieving reliable generated image detection. In this work, we propose a hierarchical retrospection refinement (HRR) framework, in which a multi-scale style retrospection module is introduced to encourage the model to generate detailed and realistic multi-scale representations, thereby mitigating the learning bias introduced by dataset styles and generative models. Additionally, we design a feature refinement module to reduce the impact of redundant features on learning, capture the intrinsic structure and patterns of the data, and enhance the model’s generalization ability. Experimental results demonstrate that our method effectively enhances the detection capability for both generated and real images. By properly selecting hyperparameters, our HRR consistently achieves state-of-the-art (SOTA) performance across three different datasets. Furthermore, qualitative visual analysis illustrates how our approach identifies subtle differences at a fine-grained level.

\vspace{-0.2cm}

\bibliographystyle{IEEEtran}
\bibliography{ref}

\vspace{-1.1cm}

\clearpage
\appendix 
%\begin{appendices}
\noindent\textit{A. Optimization Algorithm of AFR}\label{opt}

% Firstly, the objective function (\ref{alphaz}) is converted into a weighted least squares problem via the half-quadratic (HQ) optimization \cite{Nikolova05}. Then, the ADMM strategy \cite{Boyd11} is employed to solve the transformed problem.

Based on the convex optimization theory in \cite{Rockafellar97}, we define a convex function $g(b) = -b\log(-b)+b$, where $b<0$. According to the conjugate function theory in \cite{Boyd04}, we obtain
\begin{small}\begin{equation}\label{congb}
   %\exp\Big(-\frac{(1-yf(x))^2}{2\sigma^2}\Big)=\sup\limits_b \Big(b\cdot\frac{(1-yf(x))^2}{2\sigma^2}-g(b) \Big)
   {\rm{exp}}\Big(-\frac{(y-f(x))^2}{\sigma^2}\Big)=\sup\limits_{b<0} \Big(b\cdot\frac{(y-f(x))^2}{\sigma^2}-g(b) \Big),
\end{equation}\end{small}
and the supremum can be achieved at $b=-{\rm{exp}}(-\frac{(y-f(x))^2}{\sigma^2})$.

The objective function in (\ref{alphaz}) equals to
\begin{small}\begin{eqnarray*}
\max\limits_{\alpha_{j}\in\mathbb{R}^n}\frac{1}{n}\sum^{n}_{i=1} \exp\Big(-\frac{(y_i-f(x_i))^{2}}{\sigma^{2}} \Big)- \frac{\lambda}{\beta} \sum^{p}_{j=1} w_{j}\|\alpha_{j} \|_q.
\end{eqnarray*}\end{small}
Then, from (\ref{congb}), the above formulation can be rewritten as
\begin{small}\begin{eqnarray}\label{fobj}
\mathcal{R}(\alpha,b)=\max\limits_{\alpha_{j}\in\mathbb{R}^n,b\in\mathbb{R}^n}\frac{1}{n}\sum^{n}_{i=1}\Big(b_i\cdot  \frac{(y_i-f(x_i))^2}{\sigma^2}-g(b_i)\Big) ~~~ \nonumber \\
- \frac{\lambda}{\beta}\sum^{p}_{j=1} w_{j}\|\alpha_{j} \|_q,
\end{eqnarray}\end{small}
where $b=(b_1,\ldots,b_n)^T<0$ is an auxiliary vector.

Now, we use the alternating optimization method to optimize (\ref{fobj}). To be specific, given $\alpha_{j}$, we optimize over $b$, and then given $b$,  we optimize over $\alpha_{j}$.

Firstly, suppose that $\alpha_{j}$ is given, (\ref{fobj}) equals to
\begin{small}\begin{equation}\label{objB}
\max\limits_{b\in\mathbb{R}^n}\frac{1}{n}\sum^{n}_{i=1} b_i  \frac{(y_i-f(x_i))^2}{\sigma^2}-g(b_i),
\end{equation}\end{small}
where $b_i \frac{(y_i-f(x_i))^2}{\sigma^2}-g(b_i)$ are independent functions with respect to $b_i$. Hence,  we can get the solutions for (\ref{objB}) :
\begin{small}\begin{equation}\label{SoluB}
 b_i = -{\rm{exp}}\Big(-\frac{(y_i-f(x_i))^2}{\sigma^2}\Big),~~ i = 1,...,n.
\end{equation}\end{small}

Secondly, after obtaining $b$, $\alpha$ can be updated by solving the following problem:
\begin{eqnarray}\label{solution_alpha}
	\max\limits_{\alpha_{j}\in\mathbb{R}^n}\frac{1}{n}\sum^{n}_{i=1} b_i\Big(y_i-\sum^{p}_{i=1}\Psi_{ji}^T\alpha_{j}\Big)^2 - \frac{\lambda\sigma^2}{\beta} \sum^{p}_{j=1} w_{j}\|\alpha_{j} \|_q.
\end{eqnarray}
The alternation maximization of $\alpha$ and $b$ satisfies $\mathcal{R}(\alpha_t,b_t)\leq\mathcal{R}(\alpha_t,b_{t+1})\leq\mathcal{R}(\alpha_{t+1},b_{t+1})$, where $t$ denotes the $t$-th iteration. Finally, $\{\mathcal{R}(\alpha_t,b_t), t=1,2,...\}$ converges\cite{Nikolova05}.

%It is obvious that the estimation (\ref{solution_alpha}) can be regarded as a weighted least squares problem. 
Let $\alpha=({\alpha_{1}}^T,\dots,{\alpha_{p}}^T)^T\in\mathbb{R}^{dp}$, $\Phi=(\Phi_{1},\dots,\Phi_{p})\in\mathbb{R}^{n\times dp}$ and $\mathbf{Y} = (y_1,\dots,y_n)^{T}\in\mathbb{R}^{n}$. Then, the problem in (\ref{solution_alpha}) can be reformulated as
\begin{small}\begin{eqnarray}\label{solution_New}
	\min\limits_{\alpha}(\mathbf{Y}-\Phi\alpha)^{T}{\rm diag}(-b)(\mathbf{Y}-\Phi\alpha)+ \frac{\lambda\sigma^2}{\beta} \sum^{p}_{j=1} w_{j}\|\alpha_{j} \|_q,
\end{eqnarray}\end{small}
where $\rm diag(\cdot)$ is used to convert a vector to a diagonal matrix.

Denote a relax variable $\vartheta$, where  $\vartheta =({\vartheta_{1}}^T,\dots,{\vartheta_{p}}^T)^T\in\mathbb{R}^{dp}$, $\vartheta_{j}=(\vartheta_{j1},\dots,\vartheta_{jd})^T\in\mathbb{R}^d$. (\ref{solution_New}) can be translated into
\begin{small}
\begin{align}\label{solution_ALM}
%\begin{aligned}
&\min\limits_{\alpha,\vartheta}(\mathbf{Y}-\Phi\alpha)^{T}{\rm diag}(-b)(\mathbf{Y}-\Phi\alpha)+ \frac{\lambda\sigma^2}{\beta} \sum^{p}_{j=1} w_{j} \| \vartheta_{j} \|_q,\\
& s.t. \; \alpha - \vartheta = 0.\nonumber
%\end{aligned}
\end{align}
\end{small}
Hence, the scaled augmented Lagrangian function of (\ref{solution_ALM}) is
\begin{small}
\begin{eqnarray}\label{Lagrangian_func}
L(\alpha,\vartheta,\mu)= (\mathbf{Y}-\Phi\alpha)^{T}{\rm diag}(-b)(\mathbf{Y}-\Phi\alpha)+ \frac{\lambda\sigma^2}{\beta} \sum^{p}_{j=1} w_{j}\|\vartheta_{j} \|_q \nonumber\\
 +\frac{\eta}{2}\| \alpha-\vartheta +\mu\|^2_2 - \frac{\eta}{2}\|\mu\|^2_2,\nonumber
\end{eqnarray}
\end{small}
where $\eta>0$, and $\mu$ is the Lagrange multiplier. Here, this problem can be solved by the following iterative scheme:
 % More specially, when optimize one variable in the $(t+1)$-th iteration, others are kept as their latest values.
 
\noindent{\bf{(1) Update $\alpha$:}}
\begin{small}\begin{eqnarray}\label{Update_alpha}
	\alpha_{t+1}= \arg\min_{\alpha}(\mathbf{Y}-\Phi\alpha)^{T}{\rm diag}(-b)(\mathbf{Y}-\Phi\alpha) \\ \nonumber
	+  \frac{\eta}{2}\| \alpha-\vartheta_t +\mu_t\|^2_2.
\end{eqnarray}\end{small}
With fixed $\vartheta$ and $\mu$, (\ref{Update_alpha}) ia essentially a weighted ridge regression. Hence, we obtain
\begin{small}\begin{eqnarray}\label{alpha_t1}
	\alpha_{t+1}= \Big(2\Phi^T{\rm diag}(-b)\Phi+\eta \mathbf{I}\Big)^{-1}   ~~~~~~~~~~~~~~~~~~~~~~~~~\\ 
    \cdot \Big(2\Phi^T{\rm diag}(-b)\mathbf{Y} +\eta(\vartheta_t-\mu_t)  \Big). \nonumber
\end{eqnarray}\end{small}

\noindent{\bf{(2) Update $\vartheta$:}}
\begin{small}\begin{eqnarray}\label{Update_vartheta}
	\vartheta_{t+1}
	= \arg\min_{\vartheta}\frac{1}{2}\| \alpha_{t+1}-\vartheta+\mu_t\|^2_2+ \frac{\lambda\sigma^2}{\beta\eta} \sum^{p}_{j=1} w_{j}\|\vartheta_{j} \|_q.
\end{eqnarray}\end{small}

With fixed $\alpha_{t+1}$ and $\mu_t$, (\ref{Update_vartheta}) is equivalent to $p$ subproblems:
\begin{small}\begin{eqnarray*}\label{betap}
	 \arg\min_{\vartheta^{(j)} }\frac{1}{2}\| \alpha_{j,t+1}-\vartheta_{j}+\mu_{j,t}\|^2_2+ \frac{\lambda\sigma^2}{\beta\eta} \sum^{p}_{j=1} w_{j}\|\vartheta_{j} \|_q,
\end{eqnarray*}\end{small}
where $j =  1,\dots,p$.
The above subproblems can be solved by the soft thresholding operator $\mathcal{S}$ \cite{Boyd11,Chen2020}, which is
\begin{small}\begin{eqnarray}\label{betaS}
	\vartheta_{j,t+1} = \mathcal{S}_{{\lambda\sigma^2}/{\beta\eta}}(\alpha_{j,t+1}+\mu_{j,t}),\;\;j =  1,\dots,p.
\end{eqnarray}\end{small}
The soft thresholding operator $\mathcal{S}$ is defined as
\begin{small}\begin{eqnarray*}
 \mathcal{S}_{k}(a)= \left\{
    \begin{array}{ll}
          (a-k)_+-(-a-k)_+,& q = 1; \\
          (1-k/\|a\|_2)_+a ,& q = 2.
        \end{array}
\right.
\end{eqnarray*}\end{small}

%When $q=1$, the soft thresholding operator $\mathcal{S}$ is defined as
%\begin{eqnarray*}\label{S1}
%	 \mathcal{S}_{k}(a)=(a-k)_+-(-a-k)_+.
%\end{eqnarray*}
%When $q=2$, the soft thresholding operator  $\mathcal{S}$ is defined as
%\begin{eqnarray*}\label{S2}
%      \mathcal{S}_{k}(a)=(1-k/\|a\|_2)_+a.
%\end{eqnarray*}

\noindent{\bf{(3) Update $\mu$:}}
\begin{small}\begin{eqnarray}\label{mu}
	\mu_{t+1} = \mu_t+\alpha_{t+1}-\vartheta_{t+1}.
\end{eqnarray}\end{small}

Lastly, if the $(t+1)$-th iteration satisfies
\begin{small}\begin{eqnarray}\label{Stop}
    \| \alpha_{t+1}-\vartheta_{t+1}\|_\infty<\epsilon  \;{\rm{and}}\; \|\alpha_{t+1}-\alpha_{t}\|_\infty<\epsilon,
\end{eqnarray}\end{small}
stop the iteration and return $ \alpha_{t+1}$ as the final result.

\begin{algorithm}[H]
	\caption{: Optimization Algorithm of AFR.}  \label{algorithm1}
	\begin{algorithmic}
		\STATE\textbf{Input:}
		Training samples $\mathbf{z}$, $\lambda$, $\sigma$, $\epsilon$, $w_{j}$, \rm{Maxiter};
		%\STATE\textbf{Output:}
		%$\mathcal{J}_{\mathbf{z}}$ and  $\alpha_{j}_{\mathbf{z}}$; %,~j\in \mathcal{J}_{\mathbf{z}}$;
		\STATE\textbf{Initialization:} $t=0$, $\alpha_t$ via uniform distribution $U(0,1)$;
		\WHILE {not converged and $t\leq \rm{Maxiter}$}
		\STATE 1. Fix $\alpha_t$, update $b_{t+1}$ via Eq.~(\ref{SoluB});
		\STATE  2. Fix $b_{t+1}$, update $\alpha_{t+1}$ using ADMM:
		\STATE\textbf{Initialization:} $t'=0$,  $\vartheta_{t'}=\textbf{0}$, $\mu_{t'}=\textbf{0}$, $\eta = 10^{-1}$;
		\WHILE {not converged and $t'\leq \rm{Maxiter}$}
		\STATE 1) Fixed $\vartheta_{t'}$ and $\mu_{t'}$, update $\alpha_{t'+1}$ via Eq.~(\ref{alpha_t1});
		\STATE 2) Fixed $\alpha_{t'+1}$ and $\mu_{t'}$, update $\vartheta_{t'+1}$ via Eq.~(\ref{betaS});
		\STATE 3) Fixed $\alpha_{t'+1}$ and $\vartheta_{t'+1}$, update $\mu_{t'+1}$ via Eq.~\ref{mu});
		\STATE 4) Check the convergence condition in (\ref{Stop});
		\STATE 5) $t' = t'+1$;
		\ENDWHILE
		\STATE 3. $\alpha_{t+1} = \alpha_{t'}$;
		\STATE 4. Check the convergence condition: \\
      ~~~~~~~~~~~~~~~ $\mathcal{R}(\alpha_{t+1},b_{t+1})-\mathcal{R}(\alpha_{t},b_{t})\leq \epsilon$;
		\STATE 5. $t = t+1$;
		\ENDWHILE
	%	\STATE $\alpha_{\mathbf{z},j} = \alpha_{j,t},~j = 1,...,p.$
		\STATE\textbf{Output:} $\alpha_{\mathbf{z},j}=\alpha_{j,t},~j = 1,...,p.$ 
	\end{algorithmic}
\end{algorithm}
%\end{appendices}
\end{document}